\documentclass{ieeeaccess}
\usepackage{cite}
\usepackage{amsmath,amssymb,amsfonts}
\usepackage{algorithmic}
\usepackage{graphicx}
\usepackage{textcomp}
\usepackage{lscape}
\usepackage{float}

\def\BibTeX{{\rm B\kern-.05em{\sc i\kern-.025em b}\kern-.08em
T\kern-.1667em\lower.7ex\hbox{E}\kern-.125emX}}

\begin{document}

\history{2021-12-19, current version: 1.0}
\doi{None}

\title{Product Re-identification System in Fully Automated Defect Detection}

\author{Chenggui Sun and Li Bin Song}
\author{\uppercase{Chenggui Sun}\authorrefmark{1}
AND \uppercase{Li Bin Song\authorrefmark{2}}
}
\address[1]{Department of Computing Science, University of Alberta, Edmonton, Alberta, Canada T6G 2E8 (e-mail: chenggui@ualberta.ca)}
\address[2]{Department of Computing Science, University of Alberta, Edmonton, Alberta, Canada T6G 2E8 (e-mail: libin3@ualberta.ca)}

\markboth
{Author \headeretal: Preparation of Papers for IEEE TRANSACTIONS and JOURNALS}
{Author \headeretal: Preparation of Papers for IEEE TRANSACTIONS and JOURNALS}

\corresp{Corresponding author: Chenggui Sun (e-mail: chenggui@ualberta.ca) and Li Bin Song(libin3@ualberta.ca)}

\begin{abstract}
    In this work, we introduce a method and present an improved neural work to
    perform product re-identification, which is an essential core function of a
     fully automated product defect detection system. Our method is based on 
     feature distance. It's the combination of feature extraction neural 
     networks, such as VGG16, AlexNet, with an image search engine - Vearch. 
     The dataset that we used to develop product re-identification systems is 
     a water-bottle dataset that consists of 400 images of 18 types of water bottles.
     This is a small dataset, which was the biggest challenge of our work.  
     However, the combination of neural networks with Vearch shows potential to 
     tackle the product re-identification problems. Especially,
     our new neural network - AlphaAlexNet that a neural network was improved based 
     on AlexNet could improve the production identification accuracy by four percent.
     This indicates that an ideal production identification accuracy could be achieved
     when efficient feature extraction methods could be introduced and redesigned
     for image feature extractions of nearly identical products. In order to
     solve the biggest challenges caused by the small size of the dataset and 
     the difficult nature of identifying productions that have little differences
     from each other. In our future work, we propose a new roadmap to tackle
     nearly-identical production identifications: to introduce or develop new 
     algorithms that need very few images to train themselves.
\end{abstract}

\begin{IEEEkeywords}
product re-identification, feature extraction, image search, image similarity
\end{IEEEkeywords}

% \begin{keywords}
%     Enter key words or phrases in alphabetical 
%     order, separated by commas. For a list of suggested keywords, send a blank 
%     e-mail to keywords@ieee.org or visit \underline
%     {http://www.ieee.org/organizations/pubs/ani\_prod/keywrd98.txt}
% \end{keywords}

\titlepgskip=-15pt

\maketitle

\section{Introduction}

An Image search/retrieval system is a software system to find similar images to
a query image from a set of image databases. Such systems have been widely used
and studied in E-commerce, fashion industries, and person re-identification \cite{b1, b2, b3, b4}. 
Image search is a query-based image similarity matching technique. This technique
is evaluated by three major metrics: accuracy, performance on responding time,
and scalability. An image search system has two major sub-processes: query 
processes and indexing processes. In the query process, an image is given 
to the system and the system returns one or multiple images that are similar 
to the query image. The indexing process is usually a batch process. It loads 
the features of multiple pre-existing images to the database. By combining the 
query processes and indexing processes, an image search system can return 
similar images as a response to a user's request by extracting images' 
features using images' index numbers. Figure \ref{fig1} demonstrated a generalized image search system based on queries \cite{b5}. As it shows, images are preprocessed for 
further analysis at first. Then, feature vectors are extracted, and when it is
necessary, dimension reductions are performed via various steps at this step.
The obtained indexes of images are applied to perform feature matching 
by measuring the similarity distances between the query image and the store 
images in the dataset. In the end, the image(s) that matches the query image are displayed \cite{b6}.

\Figure[t!](topskip=0pt, botskip=0pt, midskip=0pt){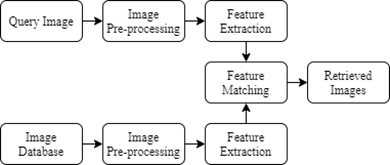}
{Scheme of Image Search Process\label{fig1}}

Image features are generally the visual descriptor of images to humans and 
include color, shape, text, and face features \cite{b7}. Color is one of the most 
used visual features of images. Color images are typically represented by 
RGB (Red, Green, Blue) space. The distributions of colors in RGB space in images
can be represented by histograms, in which one histogram is assigned to each pixel value. The quantities of histograms' bins determine the color quantization. The 
spatial relations of pixels should also be used in conjunction with color 
features. Shape feature includes edges, contours, which can be achieved by 
various methods. Image texture refers to the visual patterns, which have 
properties of homogeneity or arrangement of colors and their intensity. 
Structural and statistical representation methods have been used for texture 
representation in computer vision. 

The query subprocess of the image search isn't the research focus of this project.
Our project requires us to develop a cold-started fully automated defect 
detection system to identify a product as a new category or existing category.
If it's a new category, the images of products will be collected and processed 
to train a model to detect defects. If it's an existing category, 
then an existing category-based model will be used to perform defect detection.
The challenges of this project include:
\begin{itemize}
\item Small dataset: 400 images of 18 types of bottles;
\item Simple geometric properties of products;
\item Un-consistent colors due to non-ideal illumination conditions of the factory.
\end{itemize} 
As the sample images of water bottles in Figure \ref{fig2} show, 
the simple shape and un-consistent colors of water bottles may lead to low accuracy 
of image similarity search to identify nearly identical water bottles.

\Figure[t!](topskip=0pt, botskip=0pt, midskip=0pt){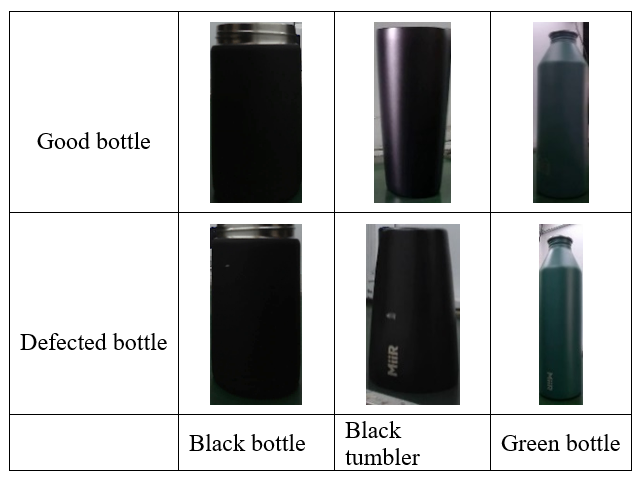}
{Six of the bottles for product re-identification algorithm development\label{fig2}}

In this work, we tested the performance of classic AlexNet classification to 
identify bottles' type as different classes and introduce a feature 
distance-based method, the combination of image feature extraction using 
neural networks and image search using an image search engine - Vearch. 
The main contributions of this work are:
\begin{itemize}
\item	Created a new dataset that consists of 400 images of 18 types of water bottles 
from monitoring videos collected from real-world production assembly lines;
\item	Examined the performance of AlexNet on object identifications using classic 
classification method when each type of water bottle was treated as one class;
\item	Introduced a feature extraction and feature-distance-based image search 
method to perform product identification. The features were extracted using pre-trained VGG16 and AlexNet, 
then feed to image search engine - Vearch to perform image search;
\item	Introduced our new AlphaAlexNet for feature extraction. 
AlphaAlexNet is an improved model with more channels that was developed on top of AlexNet.
The deployment of AlphaAlexNet could improve the image search accuracy of Vearch
by four percent than the image search accuracy of the combination of Vearch and AlexNet. 
\end{itemize} 

\section{Literature Review}
For a fully automated real-time production recognition system, the main tasks 
are to extract image frames, remove background and segment objects, 
build an image search/retrieval engine and extract features. 
There are many excellent articles in background removal, object segmentation, 
image search engine and image feature extractions for applications in facial 
recognition, gesture recognition, medical, fashion, e-commerce and 
manufacturing industries \cite{b8,b9,b10,b11,b12,b13,b14,b15,b16,b17,b18,b19,b20, b21,b22,b23,b24,b25,b26,b27,b28,b29,b30,b31,b32,b33,b34,b35,b36,b37,b38,b39,b40,b41,b42,b43,b44,b45,b46,b47,b48,b49,b50,b51,b52,b53,b54}. In this review, we will introduce and 
discuss the progress and some existing research in these areas.

\subsection*{Background Removal and Image Retrieval}
One of the challenges is that an image background is complex, and some colors are very close. In order to reduce the impact on the later step of image processing, there are a few options we can do about it. One option is to do background removal before image processing (e.g., image search and image indexing). Many pieces of research have been done on image background removal. One of the approaches is to do background subtraction based on multiple frames. Because each frame has a similar background and foreground object, the subtraction can be done. And it is used to remove the background. Zhao et al. \cite{b28} proposed a new Arithmetic Distribution Neural Network (ADNN), which can be trained for background removal tasks. This pager also analyzes traditional background removal approaches such as using the Gaussian mixture model, unsupervised Bayesian classifier. These approaches have significant limitations. Because of the complexity and diversity of all potential images, these approaches could not generate a perfect classification of pixels for background removal tasks. Other methods are built on top of Deep Learning Neural Networks, and recent research and work have been done to improve the result. The paper proposed ADNN is one of the approaches. The ADNN architecture uses the distributed neural network for background subtraction. The input is histograms of the current image and referenced histograms as counterparts from the referenced image. These inputs are passed into the arithmetic distribution layers. And a classification NN is then attached for the classification of background and foreground for pixels labels. Based on the paper, the ADNN shows promised performance and good accuracy and can be applied in real-life applications.

Qin et al. proposed a new network for object detection \cite{b27}. While once the object(foreground) is detected, the attractive object can be taken out from the background. In the paper, the team designed a deep neural network called U2-Net for object detection. The research analyzes the existing popular NN used as Salient Object Detection (SOD), such as AlexNet, VGG, ResNet, DenseNet, etc. The study reveals the limitation of these NN that all these networks are originally designed for classification. The features extracted include both foreground objects and also background information. While it is good enough for classification but there is also a limitation for saliency detection. Another boundary from the research shows that these deep network models require higher computer RAM and compute power to run, limiting the use cases. For this reason, one of the techniques is to reduce input image resolution in order to reduce RAM and compute cost. This limits the output resolution. The research team designed a new network called U2-Net. The U2-Net maintains higher image quality with low RAM usage and low CPU compute costs. The critical design of U2-Net is a two-level nested U-structure without using any pre-trained NN classification models. The U2-Net can be trained from scratch based on use cases without depending ImageNet. The U2-Net can go deeper(nested) for processing high resolution without generating extreme RAM usage and computation costs. The U2-Net can be as small as 4.7MB and up to 176.3 MB with good results against the SOTA models at 40 FPS. The U2-Net team also open source their implementation code. This code is indeed a significant contribution to this subject. A bench of real-life applications has been made on the top of U2-Net. Such as background removal app, portrait generation, etc. The U2-Net has an enormous potential that can help us with image background removal tasks.

Once we have a good image input, the next step is to identify the bottle from the image. Because of the client's requirements, any supervised learning model does not fit the needs. We are looking for a solution that can identify the bottle either is new or existing being processed before. The reason behind this is that the factory can produce a new type of bottle at any time. The system must be able to adapt to the new bottle and prepare the system for defect detection. In the object identification research area, a lot of work has been done on person identification and fashion recommendation systems. The work on those areas could not apply to this project directly because the bottle's image is so similar and has fewer features than human and fashion wearing. But the approaches are very inspiring. Most of the methods are built on top of Convolutional Neural Networks (CNNs). But in recent years, the transformers technique has been widely used in the Nature Language Processing field and gained significant success. Vision Transformer (ViT) starts to be used in image classification and identification fields. Wang et al. \cite{b26} proposed a new Pyramid Vision Transformer (PVT) method and overcame some ViT low-resolution output because of costs on computation and RAM. In their experiment, the PVT method shows promising results on object detection and semantic and instance segmentation tasks with better performance. Sharma et al. highlight their work on top of these ViT and PVT researches \cite{b25}. They proposed a Locally Aware Transformer (LA-Transformer) method and used it on person re-identification (Re-ID). The person Re-ID problem is different from the classification problem. It needs to detect a person and also be able to identify the same person from images coming from various sources (e.g., Security Cameras have no relation to each other). The work done by this research gives excellent results on high performance and high accuracy on person Re-ID problem. Person Re-ID research result inspires our project, that LA-Transformer can potentially apply on the issue of fewer features product Re-ID problem.

Once we have a method to capture the object(bottle)'s features, we need a quick way to search the image from the existing data (bottles already known). This work involves two pieces, indexing and matching. In our project use cases, there is a crucial requirement of real-time processing. This requires the solution to be able to run at low latency with high accuracy. Facebook AI team does the recent successful research and created a library(faiss) based on the study (code on github.https://github.com/facebookresearch/faiss) \cite{b24}. They proposed combining graph traversal and compact representations to accomplish the similarity search tasks. This approach is proven to use less memory and lower computation costs. In a single machine, it can handle billions of image indexing and searching.  Because this study is so successful that JD.com had another research to apply this method to E-commerce \cite{b23}. JD.com use image search to provide similar production recommendations on top of over 100 billion product images. In JD.com's study, they proposed a high-efficiency, scalable distributed system for efficient similarity search architecture. They also open source the implementation of the research code(https://github.com/vearch/vearch). The code can be either reused or modified to fit new use cases. The work from JD.com is a big help for image similarity search. Their code can be reused for image indexing.

Jeff Johnson et al. \cite{b55} proposed a design of better-utilizing GPUs to perform similarity searches. In their paper, they proposed not only a k-selection algorithm that can operate in fast register memory and be fusible with other kernels as a result of its flexibility but also a near-optimal algorithmic layout for exact and approximate k-nearest neighbor search. They applied their design for brute-force, approximate and compressed-domain search based on product quantization in different similarity search scenarios. Their algorithm and algorithmic layout enable better utilization of GPU computation, which outperformed previous art by a large margin. F. Li et al. \cite{b4} introduced a solution for e-commerce platforms to classify and search productions based on images. They trained neural networks to conduct image classification by using transfer learning with pre-trained models such as VGG19 and using autoencoders and cosine similarity to search similar images

\subsection*{Feature Extraction}
There are many approaches to extract features, including low-level local features (such as edges and corners), and global features \cite{b56}. This literature review will focus on review works that utilized deep learning based-methods to extract image features. Deep learning is essentially a neural network with three or more layers \cite{b57}, which can be stacked. The output of the previous layer is the input of the next layer, and these layers act as feature extractors. A deep learning model is an end-to-end learning model, its fully trained layers can perform much better feature extractions\cite{b58}. The most popular deep learning networks that have been used in feature extraction are AlexNet, VGGNet, ResNet and U-net \cite{b47,b48,b49}, \cite{b51}.

AlexNet consists of three types of layers: five convolutional layers, three fully-connected layers with a final 1000-way softmax, max-pooling layers that follow some of the convolutional layers. Its 60 million parameters and 650,000 neurons make overfitting become a concern. In order to reduce overfitting in the fully-connected layers, AlexNet employed “dropout” as regularization \cite{b49}.  Lv et al. \cite{b35} proposed DMS-Robust AlexNet, a neural network designed based on AlexNet architecture, to recognize maize leaf disease. Their network combined dilated convolution and multi-scale convolution to improve its capability to extract features, used batch normalization to prevent over-fitting, and PRelu activation function and Adabound optimizer to improve both convergence and accuracy. Wang and Han \cite{b32} proposed a modified AlexNet convolutional neural network for face feature point detection. This modified network has 4 convolution layers and 3 fully connected layers. It removes one layer of convolutional layers of traditional networks and adds the Batch-Normalization layers. The model's input is three sub-images of a face image, which is divided by the model, overlap with each other and have a color channel of their own. The model output the coordinates of the face feature points. Yuan and Zhang \cite{b31} used AlexNet and Caffer framework to extract features for image retrieval. The Inria Holidays and Oxford Buildings datasets were used for their experiments, which revealed that the fusion feature could contribute to the improvement of image retrieval. Besides those works that we discussed above, there are many AlexNet's applications in extracting features for hand gestures, remote sensing images, image registration and bearing fault diagnosis as well \cite{b33, b34,b36,b37,b38}.

VGGNet addresses the very important aspect of ConvNet architecture design - depth. It pushes the depth of the network to 16-19 weight layers by adding more (3 × 3) convolution filters in all layers, resulting in significantly more accurate ConvNet architectures \cite{b51}. Khaireddin1 and Chen \cite{b42} adopted the VGGNet architecture to perform facial emotion recognition on the FER2013 dataset. Their network has 4 convolutional stages that consist of two convolutional blocks and a max-pooling layer individually, and 3 fully connected layers. The convolutional stages extract features for training the fully connected layers to classify the inputs. Their model achieves state-of-the-art single-network accuracy of 73.28\% on FER2013 by tuning the model and its hyperparameters. Zhou et ai. [41] proposed to combine different granularity features from the block1, block2, block3, block4, and block5 in VGG to construct a new network architecture. A local fully connected layer was added after each block to reduce the dimensionality of the features. The combined five different granularity features are fed to the first of three global fully connected layers as input. By doing this, the information flows from a lower layer directly to a fully connected layer and the feature reuse can be increased. Their network architecture also reduces the number of parameters by removing some neurons in two global fully connected layers. This architecture was examined by using CIFAR-10 and MNIST datasets and achieved better performance than traditional VGGs. There are more works that use or adapt VGGNet to extract features for various applications \cite{b39,b40,b58}.

ResNet introduces a deep residual learning framework to address the degradation problem and shortcut connections to simply perform identity mapping, which adds neither extra parameter nor computational complexity. It's easy to optimize and gain the increases of accuracy from greatly increased depth\cite{b48}. Corbishley et al. \cite{b46} utilized the ResNet-152 CNN with super-fine attributes to recognize human attributes from surveillance video for person re-identification. After re-annotating gender, age and ethnicity of images from an amalgamation of 10 re-id datasets - PETA, their work performed significantly better to retrieval images than conventional binary labels did: a 11.2 and 14.8 percent mAP improvement for gender and age, further surpassed by ethnicity. Kwok, S. \cite{b45} presented a framework that made use of Inception-Resnet-v2 to classify breast cancer whole slide images into regions of: normal tissue, benign lesion, in-situ carcinoma and invasive carcinoma. His work won first place in ICIAR 2018 Grand Challenge on Breast Cancer Histology Images. Chen et al. \cite{b44} proposed a modularized Dual Path Network (DPN), which consists of Residual Network (ResNet) and Densely Convolutional Network (DenseNet), for image classification. This DPN architecture can reuse and explore new features with ResNet and DenseNet. It performed better than DenseNet and ResNet alone on the ImagNet-1k dataset, the PASCAL VOC detection dataset, and the PASCAL VOC. Huang et al. \cite{b43} used this ResNet-based DPN architecture to extract features and classify breast cancer images and achieved great results.

There are other great neural networks and modified networks based on these architectures used for feature extraction, such as U-Net \cite{b47}, GoogleNet \cite{b50} and MobileNets \cite{b30}. U-Net \cite{b47} has 23 convolutional layers that can be divided into a contracting stage for down-sampling and a symmetric expanding stage for up-sampling. The contracting path has two unpadded 3x3 convolutions, a ReLU and a 2x2 max pooling layer. The expanding path performs the up-sampling of the feature map and reduces the number of feature channels by half, and concatenates the corresponding feature map cropped from the contracting path. After the expanding path, a 64-component extracted feature vector is fed to the final 1x1 convolution layer for classifications. GoogleNet \cite{b50} has 22 layers and its name refers to the incarnation of the Inception module. MobileNets \cite{b30} are a class of models that can build lightweight neural networks for applications in mobile and embedded vision using depthwise separable convolutions. Its depthwise separable convolutions factorize a standard convolution into a depthwise convolution that applies a single filter to each input channel, and a 1x1 pointwise convolution that combines the outputs. Besides convolutional neural networks, there are also other networks based on Recurrent Neural Networks (RNN) and long short-term memory (LSTM) for extracting deep learning features \cite{b58}, which is not the focus of this project.

The above networks and articles that we discuss mainly focus on extracting features from images. However, the feature extraction of real-time videos is essential for releasing more potentials for fully automated defect detections. Shi et al. \cite{b29} presented the first convolutional neural network that can perform real-time super-resolution of 1080p videos on a single GPU. The feature maps are extracted in the low-resolution space, then upscaled to high-resolution output by an efficient sub-pixel convolution layer. Their approach was evaluated using publicly available datasets and performed significantly better and faster than other existing CNN-based methods.

\section{Proposed Methods}
In this project, two methods were used to perform bottle identification. Our first method is to utilize classic AlexNet to perform four types of bottles classifications to identify bottles. The first method is our baseline work. Our second method is to employ the combination of Vearch and VGG16 and AlexNet to perform the image similarity search. 

\subsection*{AlexNet}
We applied transfer learning by using pre-trained AlexNet and resigning the last layer to perform the four classes bottle classification. The architecture of AlexNet contains five convolutional layers and three fully connected layers. ReLU (Rectified Linear Unit) was used as the activation functions and MaxPooling with a stride of three was used in the pooling layers.  Dropout layers were applied to reduce overfitting. The classifier consists of seven layers: 1) two dropout layers; 2) each of the dropout layers is followed by one fully connected layer and one ReLU layer; 3) the last layer is a fully connected resigned layer with output to be set as the number of tested classes.

\subsection*{Feature Distance Based Image Similarity Search}
Our feature-distance-based image similarity search works as follows:
\begin{itemize}
    \item	Use neural networks to extract features;
    \item	Store extracted features in Vearch;
    \item	Send query/test images from dataset or cameras to feature extraction models;
    \item	Send extracted features to Vearch to perform an image search.
\end{itemize}

The core task of our works is to develop an efficient neural network for feature extraction. The feature extraction model that we tested and developed are listed as follows:
\begin{itemize}
    \item	VGG16: the pre-trained VGG16 is used to extract features from our input image.
    \item	AlexNet: the pre-trained AlexNet is used to extract features from our input image
    \item   AlphaAlexNet: six extra alpha channels are added to basic AlexNet to help the model to put more weight on shapes and colors. The definition of alpha follows the convention of the image alpha channels. 
\end{itemize}

\Figure[t!](topskip=0pt, botskip=0pt, midskip=0pt){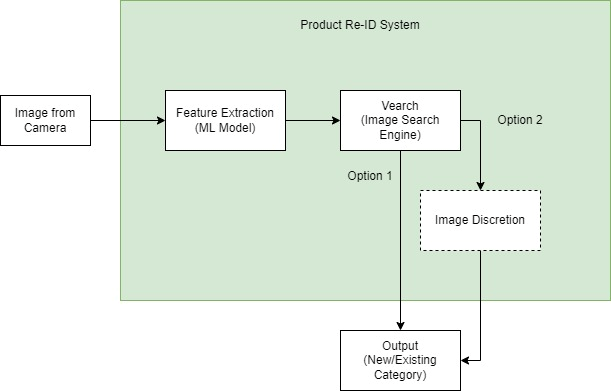}
{The design layout of bottle re-identification (query) system\label{fig3}} 

Vearch, a visual search system developed by JD E-commerce Platform, is used as our image search similarity engine. VGG16, AlexNet are two classic models for object classifications. In this work, they are used to extract features. AlphaAlexNet is integrated into Vearch search solution. After features are extracted, they are submitted to Veach for image similarity matches.

Vearch consists of two subsystems to perform indexing and search \cite{b23}. Indexing is the core subsystem of Vearch and stores extracted image features as image databases. In the original Vearch system, the indexes for all the images are periodically built. The product update events, which include the addition, deletion, and modification of a product image, can trigger Vearch to update the indexes immediately. In our design, if a new image is detected, it will be sent to the indexing process and added to the image database.

Another subsystem of Vearch is the search subsystem, which has three key components: Blender, Broker, and Searcher. When a query from a user is forwarded to one of the blenders. The blender will send the query to each broker to contact a subset of searchers to search similar images from a partition of the entire image set in parallel. The top k most similar images will be sent back to the requesting broker, which combines the results from its searchers then sends them to the blender. The combined results will be ranked by the blender and be returned to the user. Vearch's three-level architecture makes it scalable to different levels of tasks of image indexing and searches. The details of this architecture are shown in figure \ref{fig4}.

\Figure[h!](topskip=0pt, botskip=0pt, midskip=0pt){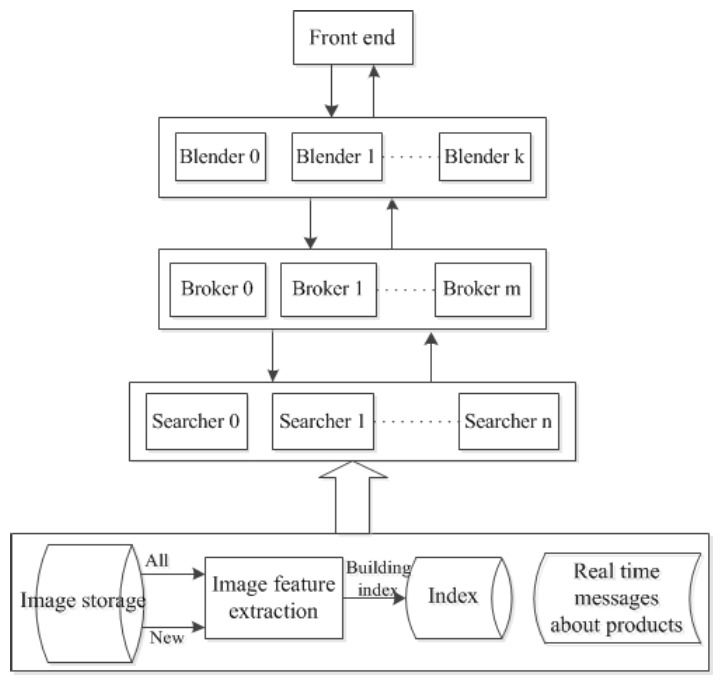}
{Architecture of Vearch System\cite{b23}\label{fig4}}

\section{Experiments}
In this section, we introduce the dataset that we used to evaluate the proposed methods and models.

We obtained 18 classes of water bottle images from production surveillance videos of a factory with the support of our client - Zerobox Inc. In order to reduce the complexity of experiments, 17 classes of bottles didn't have logos on them, except one class. Because our dataset has only about 400 images, we also tried to test our models and methods with CIFAR10 and CIFAR100. Due to the low computing performance of our computing power, these tests with CIFAR100 weren't completed. Because the size of the CIFAR10 is also small, the test on CIFA10 did not produce any meaningful results that we were expecting. Therefore, these results and implementations are not included here.

The images in figure \ref{fig5} demonstrate the 18 types of water bottles used in our research. These images only illustrate the colors and shape of the bottles, not the ratio of their actual size for visual purposes. The production backgrounds of these bottles have been removed in our pre-process stage.

\begin{table}[t]
    \centering
    \begin{tabular}{|c|c|c|c|}
    \hline
     &  AlexNet2bottle & AlexNet4bottle & AlexNet4bottle  \\
    \hline
       Dataset details & 151 back bottles & 151 back bottles & 24 black bottles\\
       & 44 white bottles & 44 white bottles & 44 white bottles \\
       & & 35 silver bottles & 35 silver bottles \\
       & & 24 blacktumbler & 24 blacktumbler \\
    \hline
    \end{tabular}
    \caption{Details of datasets for classic AlexNet classifications}
    \label{tab:table_1}
\end{table}

\section{Results and Discussion}

As what has been mentioned in the proposed methods, we investigated the image similarity match using two methods: classic Neural Network - AlexNet and feature distance-based image search-Vearch. The results of the first method were used as a baseline result to demonstrate the performance of Neural Networks due to the challenges resulting from the limited size of dataset and image quality, which motivated us to develop more efficient solutions, such as feature-distance-based image similarity match method to identify the objects in the image.

\subsubsection*{Performance of AlexNet on Similar Object Classification}
\begin{figure}[h!]
    \centering
    \includegraphics[width=0.3\textwidth]{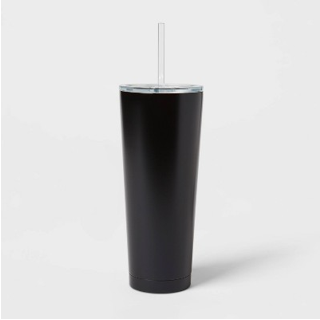}
    \caption{Black tumbler bottle}
    \label{fig7}
\end{figure}

\begin{figure}[h!]
    \centering
    \includegraphics[width=0.3\textwidth]{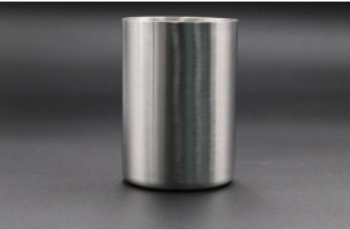}
    \caption{Silver water cup}
    \label{fig8}
\end{figure}

The train/test datasets were split at an 80:20 ratio. Detailed information about the datasets can be found in table \ref{tab:table_1} of the experiment section. Because the datasets are small, it was found that 10 epochs were enough to train the model and test the performance of trained models. The experiments on the number of epochs didn't reveal much valuable information about object identification, hence those results won't be discussed here. The overall classification accuracy and loss of our experiment on AlexNet are demonstrated in figure \ref{fig6}. These figures show that it's easy to train the model because the datasets are small. However, because the datasets are too small, although the accuracies of the train and test are good, the models weren't able to correctly identify bottles with a little bit of variation in shape or color. The black tumbler bottle shown in figure \ref{fig7} was predicted as a black bottle, the silver water cup shown in figure \ref{fig8} was also identified as a black bottle, no matter the datasets were biased or not. The results in figure \ref{fig6} also show that when more classes and images were used, the classification accuracy decreased. Generally speaking, classic models, such as AlexNet, are not fit for our project task, because of the challenges for us to collect more images and the nature of this task to identify nearly-identical objects. Therefore, we adopted feature-distance based image similarity search methodology for this project.

%\Figure[h!](topskip=0pt, botskip=0pt, midskip=0pt){fig7.png}
%{Black tumbler bottle\label{fig7}} 

%\Figure[h!](topskip=0pt, botskip=0pt, midskip=0pt){fig8.png}
%{silver water cup\label{fig8}}

\onecolumn

\Figure[h](topskip=0pt, botskip=0pt, midskip=0pt){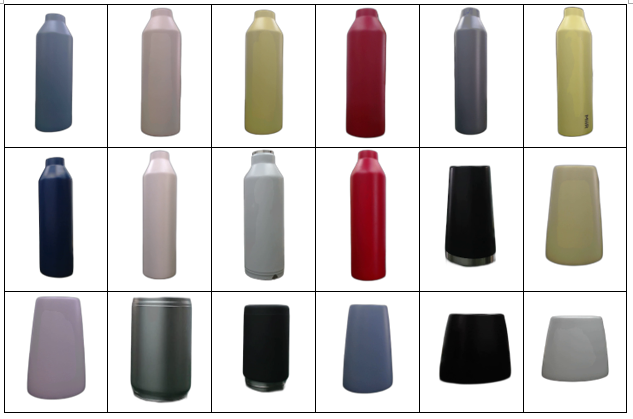}{Bottle Images Provided by Zerobox Inc. The colors of the bottles are (from right to left, from top to bottom): babyblue01, white02, beige01, red01, lavender01, yellow02, blue, white03, white, red02, black tumbler, yellow03, white01, silver, black bottles, babyblue02, black cup, white cup.\label{fig5}}

\Figure[h](topskip=0pt, botskip=0pt, midskip=0pt){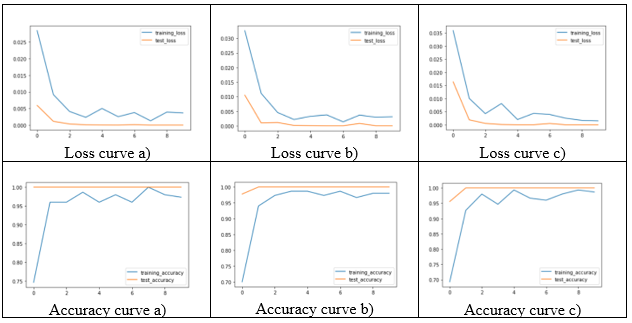}
{Loss and Accuracy curves of AlexNet over ten epochs on different datasets: 
a) two classes dataset: black and black tumbler bottles; 
b) four classes dataset: black, black tumbler bottle, silver and white-water bottles; 
c) four classes unbiased dataset with the same numbers of black and black tumbler bottles.\label{fig6}}

\twocolumn 

\subsubsection*{Performance of Feature Distance based Image Search Methods}
 Our feature-distance-based image search is completed by using pre-trained VGG16, pre-trained AlexNet and our new model - AlphaAlexNet to extract image features, storing the indexed features in the Vearch and using Vearch to perform image similarity search.
 
 The test performance of pre-trained VGG16, pre-trained AlexNet, and AlphaAlexNet on 18 bottle identifications are evaluated by confusion matrixes. The detailed results about confusion matrixes are shown in table \ref{tab:table_2} to \ref{tab:table_4}. The descriptions of the classes of bottles are listed under each table. The number after each class name indicates the minor differences in the colors of bottles. For example, white01, white02, and white03 mean that the colors of three classes of bottles are generally white, however, there are minor differences among those three white colors. From table \ref{tab:table_2} to \ref{tab:table_4}, we can tell that there 3 classes were mislabeled as other classes by the combination of pre-trained VGG16 and Vearch among 25 test bottles: 
\begin{itemize}
    \item one white01 bottle was mislabeled as a babyblue2 bottle, 
    \item one white02 was mislabeled as a babyblue01 bottle, 
    \item one white03 bottle was mislabeled as a beige01 bottle.
\end{itemize} 
 There were 4 classes that were mislabeled as other classes by the combination of pre-trained AlexNet and Vearch among 25 test bottles: 
\begin{itemize}
    \item one beige01 bottle was mislabeled as a white02 bottle, 
    \item one white01 bottle was mislabeled as a babyblue02 bottle, 
    \item one white02 bottle was mislabeled as a babyblue01 bottle,
    \item one white03 bottle was mislabeled as a white02 bottle.
\end{itemize}  
 There were 3 classes that were mislabeled as other classes by the combination of AlphaAlexNet and Vearch among 25 test bottles: 
 \begin{itemize}
    \item one black bottle was labeled as a silver bottle, 
    \item one white01 bottle was labeled as a babyblue02 bottle, 
    \item one white03 bottle was labeled as a white02 bottle.
\end{itemize}
 From the mislabeled bottles, it can be observed that it's very difficult for neural networks to distinguish bottles that have close colors and the same shapes. For example, the shape of white02 and white03 was almost the same, their colors were very close; the shape of white01 and babyblue02 was very close. Because the added six extra alpha channels put more weight on shapes and colors, the combination of AlphaAlexNet and Vearch had better performance to match similar images.

\begin{figure}[h!]
    \centering
    \includegraphics{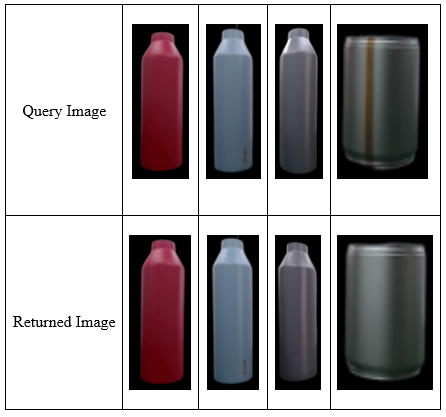}
    \caption{Real Search Examples - Correct Returns}
    \label{fig9}
\end{figure}

The overall identification accuracies of each neural networks are summarized in Table \ref{tab:table_5}. 
When the image features extracted by the pre-trained VGG16 were fed to Vearch, 
Vearch performed much better to match images than what it did when the pre-trained AlexNet was used. 
Its accuracy value is 88\% which is higher than the 84\% of pre-trained AlexNet, 
this might result from the fact that VGG has more weight layers \cite{b59}. 
When Our new AlphaAlexNet was used to extract image features, 
Vearch had an accuracy of 88\% in image similarity search. 
In the AlphaAlexNet, we added 6 extra alpha channels to the model. 
The 6 channels information comes from inverse colors and rotation of the image. 
This helped to make the model distinguish close colors and less sensitive to the 
small variations of shapes. The performance of our AlphaAlexNet with Vearch was 
also illustrated here with search samples. Figure \ref{fig9} shows the correctly returned 
images when a query image was sent to Vearch.  
Figure \ref{fig10} shows the wrong search results when a query image was sent to Vearch. 
The wrong returned search results indicate that it's a great challenge to search 
the correct images if the shape or the colors of the water bottles were close.

The results shown in Table \ref{tab:table_2} and illustrated in figure \ref{fig9} and \ref{fig10} demonstrate that
the combination of Vearch and neural networks didn't perform perfectly or reach 
our expectations in terms of search accuracy. This should have resulted from the
fact that the subtle differences among the color and shape of bottles are not well
captured by neural networks. However, our AlphaAlexNet model, which was developed
based on existing AlexNet, could improve Vearch's similarity search performance
for our task. This implies that the combination of Vearch and neural networks 
have the potential to have much better accuracies if more efficient neural networks 
could be developed to extract features or bigger datasets could be obtained to 
train the neural networks. The potential

\onecolumn

\begin{table}[]
\centering
\begin{tabular}{ccccccccccccccccccc}
No. & C1 & C2 & C3 & C4 & C5 & C6 & C7 & C8 & C9 & C10 & C11 & C12 & C13 & C14 & C15 & C16 & C17 & C18 \\
C1 & 1 & 0 & 0 & 0 & 0 & 0 & 0 & 0 & 0 & 0 & 0 & 0 & 0 & 0 & 0 & 0 & 0 & 0 \\
C2 & 0 & 1 & 0 & 0 & 0 & 0 & 0 & 0 & 0 & 0 & 0 & 0 & 0 & 0 & 0 & 0 & 0 & 0 \\
C3 & 0 & 0 & 2 & 0 & 0 & 0 & 0 & 0 & 0 & 0 & 0 & 0 & 0 & 0 & 0 & 0 & 0 & 0 \\
C4 & 0 & 0 & 0 & 4 & 0 & 0 & 0 & 0 & 0 & 0 & 0 & 0 & 0 & 0 & 0 & 0 & 0 & 0 \\
C5 & 0 & 0 & 0 & 0 & 1 & 0 & 0 & 0 & 0 & 0 & 0 & 0 & 0 & 0 & 0 & 0 & 0 & 0 \\
C6 & 0 & 0 & 0 & 0 & 0 & 1 & 0 & 0 & 0 & 0 & 0 & 0 & 0 & 0 & 0 & 0 & 0 & 0 \\
C7 & 0 & 0 & 0 & 0 & 0 & 0 & 1 & 0 & 0 & 0 & 0 & 0 & 0 & 0 & 0 & 0 & 0 & 0 \\
C8 & 0 & 0 & 0 & 0 & 0 & 0 & 0 & 1 & 0 & 0 & 0 & 0 & 0 & 0 & 0 & 0 & 0 & 0 \\
C9 & 0 & 0 & 0 & 0 & 0 & 0 & 0 & 0 & 2 & 0 & 0 & 0 & 0 & 0 & 0 & 0 & 0 & 0 \\
C10 & 0 & 0 & 0 & 0 & 0 & 0 & 0 & 0 & 0 & 2 & 0 & 0 & 0 & 0 & 0 & 0 & 0 & 0 \\
C11 & 0 & 0 & 0 & 0 & 0 & 0 & 0 & 0 & 0 & 0 & 1 & 0 & 0 & 0 & 0 & 0 & 0 & 0 \\
C12 & 0 & 0 & 0 & 0 & 0 & 0 & 0 & 0 & 0 & 0 & 0 & 2 & 0 & 0 & 0 & 0 & 0 & 0 \\
C13 & 0 & 1 & 0 & 0 & 0 & 0 & 0 & 0 & 0 & 0 & 0 & 0 & 0 & 0 & 0 & 0 & 0 & 0 \\
C14 & 1 & 0 & 0 & 0 & 0 & 0 & 0 & 0 & 0 & 0 & 0 & 0 & 0 & 0 & 0 & 0 & 0 & 0 \\
C15 & 0 & 0 & 1 & 0 & 0 & 0 & 0 & 0 & 0 & 0 & 0 & 0 & 0 & 0 & 0 & 0 & 0 & 0 \\
C16 & 0 & 0 & 0 & 0 & 0 & 0 & 0 & 0 & 0 & 0 & 0 & 0 & 0 & 0 & 0 & 1 & 0 & 0 \\
C17 & 0 & 0 & 0 & 0 & 0 & 0 & 0 & 0 & 0 & 0 & 0 & 0 & 0 & 0 & 0 & 0 & 1 & 0 \\
C18 & 0 & 0 & 0 & 0 & 0 & 0 & 0 & 0 & 0 & 0 & 0 & 0 & 0 & 0 & 0 & 0 & 0 & 1
\end{tabular}
\caption{Confusion Matrix of Image Identification: Pre-trained VGG16 + Vearch.
Classes C01 - C18:
1 Babyblue01, 2 Babyblue02, 3 Beige01, 4 Black bottle, 5 Black cup, 6 Black tumbler, 7 blue, 8 lavender01, 9 red01, 10 red02, 11 silver, 12 white, 13 white01, 14 white02, 15 white03, 16 white cup, 17 yellow02, 18 yellow03}
\label{tab:table_2}
\end{table}

\begin{table}[]
\centering
\begin{tabular}{ccccccccccccccccccc}
No. & C1 & C2 & C3 & C4 & C5 & C6 & C7 & C8 & C9 & C10 & C11 & C12 & C13 & C14 & C15 & C16 & C17 & C18 \\
C1 & 1 & 0 & 0 & 0 & 0 & 0 & 0 & 0 & 0 & 0 & 0 & 0 & 0 & 0 & 0 & 0 & 0 & 0 \\
C2 & 0 & 1 & 0 & 0 & 0 & 0 & 0 & 0 & 0 & 0 & 0 & 0 & 0 & 0 & 0 & 0 & 0 & 0 \\
C3 & 0 & 0 & 1 & 0 & 0 & 0 & 0 & 0 & 0 & 0 & 0 & 0 & 0 & 1 & 0 & 0 & 0 & 0 \\
C4 & 0 & 0 & 0 & 4 & 0 & 0 & 0 & 0 & 0 & 0 & 0 & 0 & 0 & 0 & 0 & 0 & 0 & 0 \\
C5 & 0 & 0 & 0 & 0 & 1 & 0 & 0 & 0 & 0 & 0 & 0 & 0 & 0 & 0 & 0 & 0 & 0 & 0 \\
C6 & 0 & 0 & 0 & 0 & 0 & 1 & 0 & 0 & 0 & 0 & 0 & 0 & 0 & 0 & 0 & 0 & 0 & 0 \\
C7 & 0 & 0 & 0 & 0 & 0 & 0 & 1 & 0 & 0 & 0 & 0 & 0 & 0 & 0 & 0 & 0 & 0 & 0 \\
C8 & 0 & 0 & 0 & 0 & 0 & 0 & 0 & 1 & 0 & 0 & 0 & 0 & 0 & 0 & 0 & 0 & 0 & 0 \\
C9 & 0 & 0 & 0 & 0 & 0 & 0 & 0 & 0 & 2 & 0 & 0 & 0 & 0 & 0 & 0 & 0 & 0 & 0 \\
C10 & 0 & 0 & 0 & 0 & 0 & 0 & 0 & 0 & 0 & 2 & 0 & 0 & 0 & 0 & 0 & 0 & 0 & 0 \\
C11 & 0 & 0 & 0 & 0 & 0 & 0 & 0 & 0 & 0 & 0 & 1 & 0 & 0 & 0 & 0 & 0 & 0 & 0 \\
C12 & 0 & 0 & 0 & 0 & 0 & 0 & 0 & 0 & 0 & 0 & 0 & 2 & 0 & 0 & 0 & 0 & 0 & 0 \\
C13 & 0 & 1 & 0 & 0 & 0 & 0 & 0 & 0 & 0 & 0 & 0 & 0 & 0 & 0 & 0 & 0 & 0 & 0 \\
C14 & 1 & 0 & 0 & 0 & 0 & 0 & 0 & 0 & 0 & 0 & 0 & 0 & 0 & 0 & 0 & 0 & 0 & 0 \\
C15 & 0 & 0 & 0 & 0 & 0 & 0 & 0 & 0 & 0 & 0 & 0 & 0 & 0 & 1 & 0 & 0 & 0 & 0 \\
C16 & 0 & 0 & 0 & 0 & 0 & 0 & 0 & 0 & 0 & 0 & 0 & 0 & 0 & 0 & 0 & 1 & 0 & 0 \\
C17 & 0 & 0 & 0 & 0 & 0 & 0 & 0 & 0 & 0 & 0 & 0 & 0 & 0 & 0 & 0 & 0 & 1 & 0 \\
C18 & 0 & 0 & 0 & 0 & 0 & 0 & 0 & 0 & 0 & 0 & 0 & 0 & 0 & 0 & 0 & 0 & 0 & 1
\end{tabular}
\caption{Confusion Matrix of Image Identification: Pre-trained AlexNet + Vearch.
Classes C01 - C18:
1 Babyblue01, 2 Babyblue02, 3 Beige01, 4 Black bottle, 5 Black cup, 6 Black tumbler, 7 blue, 8 lavender01, 9 red01, 10 red02, 11 silver, 12 white, 13 white01, 14 white02, 15 white03, 16 white cup, 17 yellow02, 18 yellow03}
\label{tab:table_3}
\end{table}

\begin{table}[]
\centering
\begin{tabular}{ccccccccccccccccccc}
No. & C1 & C2 & C3 & C4 & C5 & C6 & C7 & C8 & C9 & C10 & C11 & C12 & C13 & C14 & C15 & C16 & C17 & C18 \\
C1 & 1 & 0 & 0 & 0 & 0 & 0 & 0 & 0 & 0 & 0 & 0 & 0 & 0 & 0 & 0 & 0 & 0 & 0 \\
C2 & 0 & 1 & 0 & 0 & 0 & 0 & 0 & 0 & 0 & 0 & 0 & 0 & 0 & 0 & 0 & 0 & 0 & 0 \\
C3 & 0 & 0 & 2 & 0 & 0 & 0 & 0 & 0 & 0 & 0 & 0 & 0 & 0 & 0 & 0 & 0 & 0 & 0 \\
C4 & 0 & 0 & 0 & 4 & 0 & 0 & 0 & 0 & 0 & 0 & 0 & 0 & 0 & 0 & 0 & 0 & 0 & 0 \\
C5 & 0 & 0 & 0 & 0 & 1 & 0 & 0 & 0 & 0 & 0 & 0 & 0 & 0 & 0 & 0 & 0 & 0 & 0 \\
C6 & 0 & 0 & 0 & 0 & 0 & 0 & 0 & 0 & 0 & 0 & 1 & 0 & 0 & 0 & 0 & 0 & 0 & 0 \\
C7 & 0 & 0 & 0 & 0 & 0 & 0 & 1 & 0 & 0 & 0 & 0 & 0 & 0 & 0 & 0 & 0 & 0 & 0 \\
C8 & 0 & 0 & 0 & 0 & 0 & 0 & 0 & 1 & 0 & 0 & 0 & 0 & 0 & 0 & 0 & 0 & 0 & 0 \\
C9 & 0 & 0 & 0 & 0 & 0 & 0 & 0 & 0 & 2 & 0 & 0 & 0 & 0 & 0 & 0 & 0 & 0 & 0 \\
C10 & 0 & 0 & 0 & 0 & 0 & 0 & 0 & 0 & 0 & 2 & 0 & 0 & 0 & 0 & 0 & 0 & 0 & 0 \\
C11 & 0 & 0 & 0 & 0 & 0 & 0 & 0 & 0 & 0 & 0 & 1 & 0 & 0 & 0 & 0 & 0 & 0 & 0 \\
C12 & 0 & 0 & 0 & 0 & 0 & 0 & 0 & 0 & 0 & 0 & 0 & 2 & 0 & 0 & 0 & 0 & 0 & 0 \\
C13 & 0 & 1 & 0 & 0 & 0 & 0 & 0 & 0 & 0 & 0 & 0 & 0 & 0 & 0 & 0 & 0 & 0 & 0 \\
C14 & 0 & 0 & 0 & 0 & 0 & 0 & 0 & 0 & 0 & 0 & 0 & 0 & 0 & 1 & 0 & 0 & 0 & 0 \\
C15 & 0 & 0 & 0 & 0 & 0 & 0 & 0 & 0 & 0 & 0 & 0 & 0 & 0 & 1 & 0 & 0 & 0 & 0 \\
C16 & 0 & 0 & 0 & 0 & 0 & 0 & 0 & 0 & 0 & 0 & 0 & 0 & 0 & 0 & 0 & 1 & 0 & 0 \\
C17 & 0 & 0 & 0 & 0 & 0 & 0 & 0 & 0 & 0 & 0 & 0 & 0 & 0 & 0 & 0 & 0 & 1 & 0 \\
C18 & 0 & 0 & 0 & 0 & 0 & 0 & 0 & 0 & 0 & 0 & 0 & 0 & 0 & 0 & 0 & 0 & 0 & 1
\end{tabular}
\caption{Confusion Matrix of Image Identification: AlphaAlexNet + Vearch.
Classes C01 - C18:
1 Babyblue01, 2 Babyblue02, 3 Beige01, 4 Black bottle, 5 Black cup, 6 Black tumbler, 7 blue, 8 lavender01, 9 red01, 10 red02, 11 silver, 12 white, 13 white01, 14 white02, 15 white03, 16 white cup, 17 yellow02, 18 yellow03}
\label{tab:table_4}
\end{table}

\onecolumn
\begin{table}[t]
\centering
\begin{tabular}{|l|l|l|l|}
\hline
Feature Extraction Model & Pre-trained VGG16 & Pre-trained AlexNet & AlphaAlexNet \\ \hline
Image Search Accuracy & \multicolumn{1}{c|}{88\%} & \multicolumn{1}{c|}{84\%} & \multicolumn{1}{c|}{88\%} \\ \hline
\end{tabular}
\caption{Image Search Accuracy of Vearch with different feature extraction methods}
\label{tab:table_5}
\end{table}

\Figure[h!](topskip=0pt, botskip=0pt, midskip=0pt){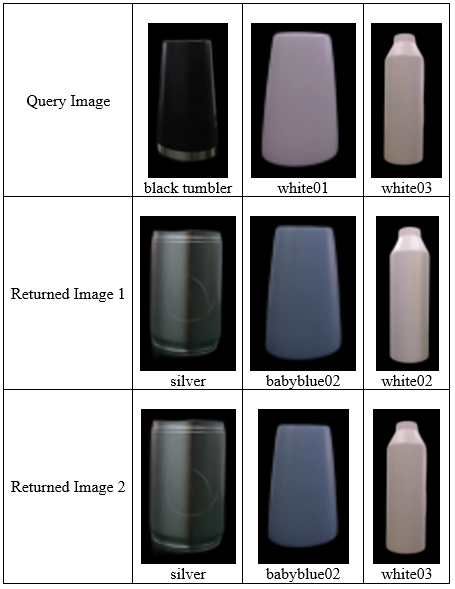}
{Real Search Examples - False Returns \label{fig10}}

\twocolumn

of Veach's commercialization in the 
product re-identification could also be supported by the work of JD e-commerce \cite{b23}.  
The combination of Vearch and neural networks has many advantages over neural
network classifications, such as AlexNet classifications. It can perform image 
search at a very quick speed, and the speed doesn't deteriorate significantly 
when dataset size and object classes increase. However, as we have discussed earlier. 
More work should be done to improve its accuracy.

\section{Ongoing Research: Siamese-AlexNet for Final Discretion}
\begin{figure}[h!]
    \centering
    \includegraphics[scale=0.7]{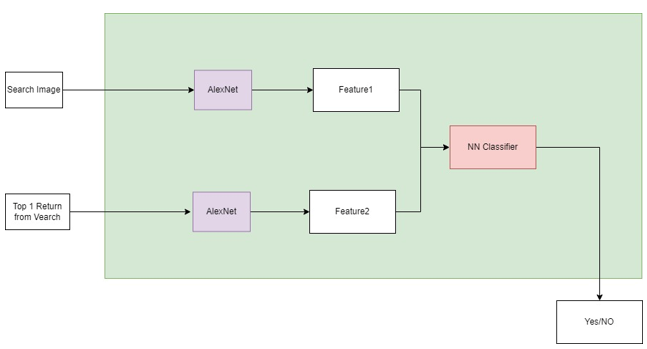}
    \caption{SimameseAlexNet Architecture}
    \label{fig11}
\end{figure}

We have been developing a method to improve Vearch's product identification algorithm. 
Currently, Vearch returns similarity search results by ranking the search results 
based on the feature distance between images. In our case, after Vearch returned 
an image of a water bottle, a decision should be made whether this query image 
is in the same category as the returned image or whether a new type of bottle 
is detected. To improve the accuracy of this decision, we designed a new model
“Siamese-Alexnet”. The model architecture is shown in figure \ref{fig11}. In training, 
an accuracy of 100\% was achieved. However, this network produced an accuracy of 50\% in real query cases.
The result isn't expected. The main reason for this should be the small size of
the training dataset. We are working on improving this network. This will also 
be part of our future work get enough images to train this model or improve model architecture for this job.

\section{Conclusion and Future Work}

In this work, we investigated solutions to tackle a nearly identical object 
re-identification problem, which means to identify an object and decide the 
category it belongs to.  A water bottle dataset consisting of 400 images of 
18 different types of bottles is used. The classic classification method was tested,
which wasn't suitable for this object re-identification problem. Then feature 
distance-based methodology, the combination of image feature extraction neural 
networks, such as AlexNet and VGG16 and image search engine - Vearch, 
were used to perform re-identification of objects. Our new model - AlphaAlexNet,
an improved model based on AlexNet, could improve the image search performance of Vearch.
This indicates that the combination of improved neural networks and Vearch 
could have great potential to tackle the object identifications problems in manufacturing. 
Based on the challenges that we were facing, our experimental results, and 
research on exiting neural networks, we concluded that we can do the following 
to develop an efficient method to re-identify objects with high accuracy in our future work: 
\begin{itemize}
    \item The combination of neural networks and Vearch has great potential. However, more efficient feature extraction neural networks should be developed with a focus on the color and shape features of objects with simple geometric and textural properties. 
    \item A new re-identification neural network should be introduced to Vearch to help Vearch to make identification decisions. This network could be the Siamese-AlexNet that we have designed or another network.
    \item Throughout the work of this project, the biggest challenge has always been lacking a high-quality image dataset of products, in our case, water bottles. Moreover, in the production settings, it's doubtful whether the increase of dataset size could improve the performance of the classic neural networks because of the nature of the production re-identifications problem - to identify nearly identical problems, which means the products in the images won't have that much different. Therefore, a new algorithm that is more suitable for production re-identifications with limited images should be introduced, just like that DifferNet can solve the surface defects detection problem with as little as 16 images. 
\end{itemize}

%\newpage

\EOD


\begin{thebibliography}{00}
    
    \bibitem{b1} M. Wieczorek, B. Rychalska, and J. Dabrowski, “On the Unreasonable Effectiveness of Centroids in Image Retrieval,” Apr. 2021, [Online]. Available: http://arxiv.org/abs/2104.13643
    
    \bibitem{b2} Z. Liu, S. Yan, P. Luo, X. Wang, and X. Tang, “Fashion Landmark Detection in the Wild,” Aug. 2016, [Online]. Available: http://arxiv.org/abs/1608.03049
    
    \bibitem{b3} S. Park, M. Shin, S. Ham, S. Choe, and Y. Kang, “Study on Fashion Image Retrieval Methods for Efficient Fashion Visual Search.”
    
    \bibitem{b4} F. Li, S. Kant, S. Araki, S. Bangera, and S. S. Shukla, “Neural Networks for Fashion Image Classification and Visual Search,” May 2020, [Online]. Available: http://arxiv.org/abs/2005.08170
    
    \bibitem{b5} J. Zhao, Z. Sihao, and Z. Jing, “Review of the sparse coding and the applications on image retrieval,” 2016. doi: 10.1109/CESYS.2016.7889904.
    
    \bibitem{b6} F. Alaei, A. Alaei, M. Blumenstein, and U. Pal, “A brief review of document image retrieval methods: Recent advances,” in Proceedings of the International Joint Conference on Neural Networks, Oct. 2016, vol. 2016-October, pp. 3500-3507. doi: 10.1109/IJCNN.2016.7727648.
    
    \bibitem{b7} M. Oussalah, “Content based image retrieval: Review of state of art and future directions,” 2008. doi: 10.1109/IPTA.2008.4743799.
    
    \bibitem{b8} I. Cheng, S. Nilufar, C. Flores-Mir, and A. Basu, “Airway Segmentation and Measurement in CT Images”.
    
    \bibitem{b9} J. Yang, M. Faraji, and A. Basu, “Robust segmentation of arterial walls in intravascular ultrasound images using Dual Path U-Net,” Ultrasonics, vol. 96, pp. 24-33, Jul. 2019, doi: 10.1016/j.ultras.2019.03.014.
    
    \bibitem{b10} Minn., IEEE Engineering in Medicine and Biology Society. Annual Conference (31st : 2009 : Minneapolis et al., “Gradient Vector Flow based Active Shape Model for Lung Field Segmentation in Chest Radiographs,” in Proceedings of the 31st Annual International Conference of the IEEE Engineering in Medicine and Biology Society, 2009, pp. 3561-3564.
    
    \bibitem{b11} M. Faraji, I. Cheng, I. Naudin, and A. Basu, “Segmentation of Arterial Walls in Intravascular Ultrasound Cross-Sectional Images Using Extremal Region Selection,” 2018. [Online]. Available: http://creativecommons.org/licenses/by-nc-nd/4.0/
    
    \bibitem{b12} M. Singh, M. Mandal, and A. Basu, “Pose Recognition using the Radon Transform,” 2005.
    
    \bibitem{b13} M. Singh, M. Mandal, and A. Basu, “Visual gesture recognition for ground air traffic control using the Radon transform,” p. 10, 2005, doi: 10.1109/IROS.2005.1545408ï.
   
    \bibitem{b14} L. Yin and A. Basu, “Integrating active face tracking with model based coding,” 1999. [Online]. Available: www.elsevier.nl/locate/patrec
   
    \bibitem{b15} M. Fiala and A. Basu, “Hough transform for feature detection in panoramic images,” 2002. [Online]. Available: www.elsevier.com/locate/patrec
   
    \bibitem{b16} R. Shen, I. Cheng, X. Li, and A. Basu, “Stereo Matching Using Random Walks,” 2008. [Online]. Available: http://vision.middlebury.edu/stereo/
   
    \bibitem{b17} L. Jansen, N. Liebrecht, S. Soltaninejad, and A. Basu, “3D Object Classification Using 2D Perspectives of Point Clouds,” in Smart Multimedia, T. McDaniel, S. Berretti, I. D. D. Curcio, and A. Basu, Eds. 2020, pp. 453-462. [Online]. Available: http://www.springer.com/series/7409
   
    \bibitem{b18} Y. Ma, G. Dong, C. Zhao, A. Basu, and Z. Wu, “Background Subtraction Based on Principal Motion for a Freely Moving Camera,” in Smart Multimedia, T. McDaniel, S. Berretti, I. D. D. Curcio, and A. Basu, Eds. 2020, pp. 67-78. [Online]. Available: http://www.springer.com/series/7409
   
    \bibitem{b19} X. Wu, X. Gao, C. Zhao, J. Wu, and A. Basu, “Background Subtraction by Difference Clustering,” in Smart Multimedia, T. McDaniel, S. Berretti, I. D. D. Curcio, and A. Basu, Eds. 2020, pp. 45-56. [Online]. Available: http://www.springer.com/series/7409
   
    \bibitem{b20} J. B. Graham-Knight et al., “Accurate Kidney Segmentation in CT Scans Using Deep Transfer Learning,” in Smart MultiMedia, T. McDaniel, S. Berretti, I. D. D. Curcio, and A. Basu, Eds. 2020, pp. 147-157. [Online]. Available: http://www.springer.com/series/7409
   
    \bibitem{b21} S. Mukherjee, G. Valenzise, and I. Cheng, “Potential of Deep Features for Opinion-Unaware, Distortion-Unaware, No-Reference Image Quality Assessment,” in Smart Multimedia, T. McDaniel, S. Berretti, I. D. D. Curcio, and A. Basu, Eds. 2020, pp. 87-95. [Online]. Available: http://www.springer.com/series/7409
   
    \bibitem{b22} G. Lugo, N. Hajari, A. Reddy, and I. Cheng, “Textureless Object Recognition Using an RGB-D Sensor,” in Smart Multimedia, T. McDaniel, S. Berretti, I. D. D. , Curcio, and A. Basu, Eds. 2020, pp. 13-27. [Online]. Available: http://www.springer.com/series/7409
   
    \bibitem{b23} J. Li et al., “The Design and Implementation of a Real Time Visual Search System on JD E-commerce Platform.”
   
    \bibitem{b24} M. Douze, A. Sablayrolles, H. Jégou, † Facebook, and A. I. Research, “Link and code: Fast indexing with graphs and compact regression codes.”
   
    \bibitem{b25} C. Sharma, S. R. Kapil, and D. Chapman, “Person Re-Identification with a Locally Aware Transformer.”
   
    \bibitem{b26} W. Wang et al., “Pyramid Vision Transformer: A Versatile Backbone for Dense Prediction without Convolutions,” Feb. 2021, [Online]. Available: http://arxiv.org/abs/2102.12122
   
    \bibitem{b27} X. Qin, Z. Zhang, C. Huang, M. Dehghan, O. R. Zaiane, and M. Jagersand, “U2-Net: Going Deeper with Nested U-Structure for Salient Object Detection,” May 2020, doi: 10.1016/j.patcog.2020.107404.
   
    \bibitem{b28} C. Zhao, K. Hu, and A. Basu, “Arithmetic Distribution Neural Network for Background Subtraction,” Apr. 2021, [Online]. Available: http://arxiv.org/abs/2104.08390
   
    \bibitem{b29} W. Shi et al., “Real-Time Single Image and Video Super-Resolution Using an Efficient Sub-Pixel Convolutional Neural Network,” Sep. 2016, [Online]. Available: http://arxiv.org/abs/1609.05158
   
    \bibitem{b30} A. G. Howard et al., “MobileNets: Efficient Convolutional Neural Networks for Mobile Vision Applications,” Apr. 2017, [Online]. Available: http://arxiv.org/abs/1704.04861
   
    \bibitem{b31} Z.-W. Yuan and J. Zhang, “Feature extraction and image retrieval based on AlexNet,” in Eighth International Conference on Digital Image Processing (ICDIP 2016), Aug. 2016, vol. 10033, p. 100330E. doi: 10.1117/12.2243849.
   
    \bibitem{b32} Wang Huai and Han Huo, “An Improved AlexNet Model with Multi-channel Input Images Processing for Human Face Feature Points Detection,” in 2020 12th International Conference on Communication Software and Networks (ICCSN 2020) , 2020, pp. 246-251.
   
    \bibitem{b33} Institute of Electrical and Electronics Engineers, “Low-cost GelSight with UV Markings: Feature Extraction of Objects Using AlexNet and Optical Flow without 3D Image Reconstruction,” in IEEE International Conference on Robotics and Automation (ICRA) , 2020, pp. 3680-3685.

    \bibitem{b34} T. Lu, F. Yu, B. Han, and J. Wang, “A generic intelligent bearing fault diagnosis system using convolutional neural networks with transfer learning,” IEEE Access, vol. 8, pp. 164807-164814, 2020, doi: 10.1109/ACCESS.2020.3022840.

    \bibitem{b35} M. Lv, G. Zhou, M. He, A. Chen, W. Zhang, and Y. Hu, “Maize Leaf Disease Identification Based on Feature Enhancement and DMS-Robust Alexnet,” IEEE Access, vol. 8, pp. 57952-57966, 2020, doi: 10.1109/ACCESS.2020.2982443.

    \bibitem{b36} K. Kavitha, B. T. Rao, A. P. Guntur, and I. B. Sandhya, “Evaluation of Distance Measures for Feature based Image Registration using AlexNet,” 2018. [Online]. Available: www.ijacsa.thesai.org

    \bibitem{b37} L. Ding, H. Li, C. Hu, W. Zhang, and S. Wang, “Alexnet feature extraction and multi-kernel learning for object-oriented classification,” in International Archives of the Photogrammetry, Remote Sensing and Spatial Information Sciences - ISPRS Archives, Apr. 2018, vol. 42, no. 3, pp. 277-281. doi: 10.5194/isprs-archives-XLII-3-277-2018.

    \bibitem{b38} Barbhuiya Abul Abbas, Karsh Ram Kumar, and Dutta Samiran, “AlexNet-CNN Based Feature Extraction and Classification of Multiclass ASL Hand Gestures,” in Proceeding of Fifth International Conference on Microelectronics, Computing and Communication Systems, 2020, pp. 77-89. [Online]. Available: http://www.springer.com/series/7818

    \bibitem{b39} X.-Y. Dai, Q.-H. Meng, W.-J. Zheng, and S.-K. Zhu, “Monocular Visual SLAM based on VGG Feature Point Extraction.”

    \bibitem{b40} Q. Zhang, “Facial expression recognition in VGG network based on LBP feature extraction,” in Proceedings - 2020 5th International Conference on Mechanical, Control and Computer Engineering, ICMCCE 2020, Dec. 2020, pp. 2089-2092. doi: 10.1109/ICMCCE51767.2020.00454.

    \bibitem{b41} Y. Zhou, H. Chang, Y. Lu, X. Lu, and R. Zhou, “Improving the Performance of VGG through Different Granularity Feature Combinations,” IEEE Access, vol. 9, pp. 26208-26220, 2021, doi: 10.1109/ACCESS.2020.3031908.

    \bibitem{b42} Y. Khaireddin and Z. Chen, “Facial Emotion Recognition: State of the Art Performance on FER2013.”

    \bibitem{b43} C.-H. Huang et al., “Automated Breast Cancer Image Classification Based on Integration of Noisy-And Model and Fully Connected Network,” in Image Analysis and Recognition, 2018, vol. 10882, pp. 923-930. doi: 10.1007/978-3-319-93000-8.

    \bibitem{b44} Y. Chen, J. Li, H. Xiao, X. Jin, S. Yan, and J. Feng, “Dual Path Networks,” Jul. 2017, [Online]. Available: http://arxiv.org/abs/1707.01629

    \bibitem{b45} S. Kwok, “Multiclass Classification of Breast Cancer in Whole-Slide Images,” in Image Analysis and Recognition, 2018, vol. 10882, pp. 931-940. doi: 10.1007/978-3-319-93000-8.

    \bibitem{b46} D. Martinho-Corbishley, M. S. Nixon, and J. N. Carter, “Super-Fine Attributes with Crowd Prototyping,” IEEE Transactions on Pattern Analysis and Machine Intelligence, vol. 41, no. 6, pp. 1486-1500, Jun. 2019, doi: 10.1109/TPAMI.2018.2836900.

    \bibitem{b47} O. Ronneberger, P. Fischer, and T. Brox, “U-Net: Convolutional Networks for Biomedical Image Segmentation,” May 2015, [Online]. Available: http://arxiv.org/abs/1505.04597

    \bibitem{b48} K. He, X. Zhang, S. Ren, and J. Sun, “Deep Residual Learning for Image Recognition,” Dec. 2015, [Online]. Available: http://arxiv.org/abs/1512.03385

    \bibitem{b49} A. Krizhevsky, I. Sutskever, and G. E. Hinton, “ImageNet Classification with Deep Convolutional Neural Networks.” [Online]. Available: http://code.google.com/p/cuda-convnet/

    \bibitem{b50} C. Szegedy et al., “Going Deeper with Convolutions.”

    \bibitem{b51} K. Simonyan and A. Zisserman, “Very Deep Convolutional Networks for Large-Scale Image Recognition,” Sep. 2014, [Online]. Available: http://arxiv.org/abs/1409.1556

    \bibitem{b52} Dharani Devi P and Thanuja. R, “Convolutional Neural Network based Deep Feature Extraction in Remote Sensing Images,” in Proceedings, International Conference on Smart Electronics and Communication (ICOSEC 2020) , 2020, pp. 441-444.

    \bibitem{b53} M. Fiala and A. Basu, “Panoramic stereo reconstruction using non-SVP optics,” 2005.

    \bibitem{b54} L. Yin and A. Basu, “Nose shape estimation and tracking for model-based coding,” 2001.

    \bibitem{b55} J. Johnson, M. Douze, and H. Jégou, “Billion-scale similarity search with GPUs,” Feb. 2017, [Online]. Available: http://arxiv.org/abs/1702.08734

    \bibitem{b56} M. S. Nixon and A. S. Aguado, “Feature Extraction and Image Processing for Computer Vision,” in Feature Extraction and Image Processing for Computer Vision, Elsevier, 2020, p. iii. doi: 10.1016/b978-0-12-814976-8.01001-0.

    \bibitem{b57} IBM, “Deep Learning,” https://www.ibm.com/cloud/learn/deep-learning.

    \bibitem{b58} C. Li et al., “A Comprehensive Review of Computer-aided Whole-slide Image Analysis: from Datasets to Feature Extraction, Segmentation, Classification, and Detection Approaches,” Feb. 2021, [Online]. Available: http://arxiv.org/abs/2102.10553

    \bibitem{b59} K. Simonyan and A. Zisserman, “Very Deep Convolutional Networks for Large-Scale Image Recognition,” Sep. 2014, [Online]. Available: http://arxiv.org/abs/1409.1556

    % \bibitem{b10} G. O. Young, ``Synthetic structure of industrial
    % plastics,'' in Plastics, vol. 3, Polymers of Hexadromicon, J. Peters,
    % Ed., 2\textsuperscript{nd} ed. New York, NY, USA: McGraw-Hill, 1964, pp. 15-64.
    % [Online]. Available:
    % \underline{http://www.bookref.com}.
    
\end{thebibliography}
\end{document}